\documentclass{article}

\PassOptionsToPackage{numbers,compress}{natbib}
\usepackage[final,dblblindworkshop]{neurips_2026}

\usepackage[utf8]{inputenc}
\usepackage[T1]{fontenc}
\usepackage{hyperref}
\usepackage{url}
\usepackage{booktabs}
\usepackage{graphicx}
\usepackage{amsfonts}
\usepackage{amsmath}
\usepackage{nicefrac}
\usepackage{microtype}
\usepackage{xcolor}

\title{Pooling Helps, Learned Weighting Hurts In-Context: Decomposing Group Attention}

\workshoptitle{Foundation Models for Temporal Systems}

\author{%
  Michael Fore, James Mason Inder, Mrishika Nair \\
  \textbf{Praneetha Vaddamanu, Sharlina Keshava} \\
  Amazon \\
  \texttt{\{mikefore, jamesind, mrishikn, lpranvad, skeshava\}@amazon.com}
}

\begin{document}

\maketitle

\begin{abstract}
Group attention, introduced by the time series forecasting model Chronos-2, attends over the variates of a group at a fixed patch index and serves both multivariate (MV) and in-context learning (ICL) forecasting. Rather than evaluating this cross-variate attention design as a whole, we ask which \emph{part} of the mechanism earns the benefit and probe its applicability to \emph{both} MV and ICL regimes. By editing the attention matrix $\alpha$ at inference we separate the two pathways a head comprises: V/O, which projects a weighted summary of the group, and Q/K, which decides the weights. Uniform pooling (V/O without any Q/K weighting) is positive on 18 of our 20 sensor-network configurations, while the learned weighting (Q/K) splits by group type: its contribution is positive or negligible for MV, but materially degrades 8 of the 10 sensor-network ICL configurations, leaving 4 of them worse than \emph{univariate} inference. Ensuring ICL groups are phase aligned removes most of the harm. Where examples cannot be aligned, forcing $\alpha$ uniform in the first block alone improves every ICL configuration we test.
\end{abstract}

\section{Introduction}
Despite the prevalence of methods using covariates in time series forecasting \citep{itransformer,softs,ccm}, univariate models remain strong baselines. The standard explanation is a capacity--robustness trade-off, where cross-variate capacity is paid for in sensitivity to drift \citep{patchtst,dlinear,hanci}. Recent evidence sharpens this: many \emph{learned} components for handling covariates provide limited value over fixed or uniform alternatives \citep{averagetime,catsattn,uniformspatial,universalredundancies}.

We examine the group attention mechanism of Chronos-2 \citep{chronos2} because it treats univariate forecasting as a first-class mode, so suppressing cross-variate flow leaves the model in a configuration it was trained for, enabling us to ablate inclusion of cross-variate information at inference time (Moirai \citep{moirai}, which flattens variates and time, offers no such separation). Additionally, group attention is advertised to serve \emph{both} multivariate forecasting (distinct related sensors measuring at the same time) \emph{and} in-context learning (windows drawn from the same or similar sensors at other times as examples). Group attention operates along the batch axis over the variates of a group at a \emph{fixed patch index}. Given there is no reason to expect in-context learning examples to be time or phase aligned, we test how this single mechanism effectively serves both use cases.

Group attention heads comprise a Q/K projection which defines the \emph{weights} in a weighted average and a V/O projection that constitutes \emph{what} gets averaged \citep{elhage}. By editing the learned Q/K weighting ($\alpha$) at inference we isolate the effect of each pathway (Figure~\ref{fig:hero}). We find uniform pooling (V/O) helps MV and ICL, but learned Q/K weighting helps only multivariate groups and is harmful for ICL. Phase-aligning ICL groups removes most of the harm. Where examples cannot be aligned, forcing $\alpha$ uniform in the first block alone with no retraining improves every in-context configuration we test, materially on 8 of 10.

\section{Method}
\label{sec:method}

\begin{figure}[t]
\centering
\includegraphics[width=\textwidth]{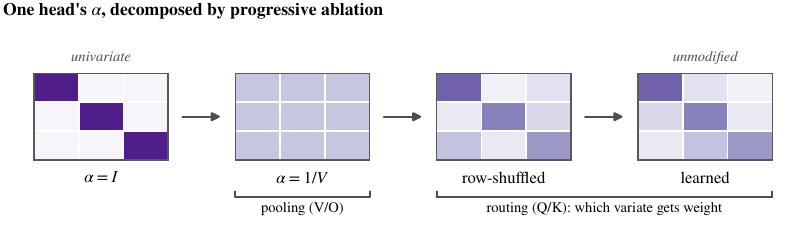}
\caption{The four $\alpha$ settings of \S\ref{sec:method}: darker cells are larger weights, and consecutive differences isolate pooling (V/O) from learned weighting (Q/K).}
\label{fig:hero}
\end{figure}

%\textbf{Decomposing the mechanism.} 
Group attention produces, per (patch index, head), a $V \times V$ matrix $\alpha$ over the variates in a group, each row a softmax distribution. One head's cross-variate update to variate $i$ is $\sum_j \alpha_{ij} W_{OV} x_j$: a weighted average of the group, projected into $i$'s representation. We use the public Chronos-2 checkpoint and replace $\alpha$ in each of the twelve encoder blocks. Four settings each progressively restore one more part of the mechanism than the last (Figure~\ref{fig:hero}). $\alpha = I$ lets no cross-variate information flow, reproducing univariate inference. $\alpha = 1/V$ is uniform \textbf{pooling}, a standard null for attention \citep{wiegreffe,hardcoded}, which shows what V/O contributes on its own. The \textbf{row-shuffled} $\alpha$ permutes each row's off-diagonal entries. This preserves every weight in a row, maintaining sharpening and entropy effects, and changes only which variate receives which weight. The learned $\alpha$ is unmodified Chronos-2 (\textbf{learned weighting}). Learned-minus-row-shuffled therefore measures whether it matters \emph{which} variate gets larger weight.

%\paragraph{Data.}
We test on five sensor networks: Atlantic/Gulf and Pacific tide gauges, USCRN climate stations, PJM load zones, and Central-European national loads. MV groups are eight stations over a shared window; ICL groups are eight windows of one randomly chosen station, starting in distinct non-overlapping slots of the record so that no window's context can contain another's horizon. Both are groupings Chronos-2 \citep{chronos2} documents: variates with shared dynamics and series with shared source. Each dataset is evaluated at two protocols (204 context and 52 horizon, and 320/96), giving 20 configurations, 10 per mode.

%\paragraph{Measurement.} 
We treat our four settings as a progressive ladder starting from a univariate baseline, reporting contributions in points of univariate MAE above the previous setting. This is scale-free across datasets; absolute errors are in Appendix~\ref{app:repair}. The unit of evaluation is a group: one forward pass over its $V$ series, scored by MAE averaged over all members and the horizon. Intervals are a 95\% paired bootstrap over groups (10,000 resamples): every condition is scored on the same groups, and we resample groups and average their per-group differences. Because any intervention yields an interval excluding zero given enough groups, we call an effect \emph{material} only when its 95\% interval lies beyond $\pm 0.1$ points. This floor is calibrated on block 5, whose learned $\alpha$ is already near-uniform: forcing it uniform moves forecasts by at most $0.081$ points, yet 14 of 20 intervals exclude zero. The row-shuffled score is the mean per-group error over eight permutations.

\section{Results}
\label{sec:results}

\subsection{Learned weighting helps multivariate groups and hurts in-context groups}
\label{sec:dissociation}

\begin{figure}[t]
\centering
\includegraphics[width=\textwidth]{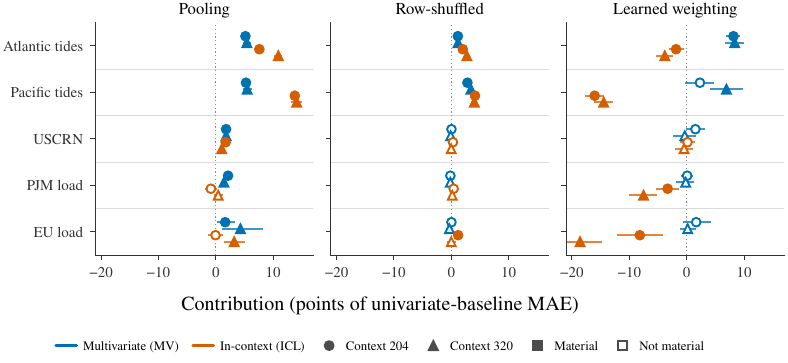}
\caption{The three contributions on five sensor networks, each as an MV and an ICL group at two protocols. Bars are 95\% paired bootstrap intervals; filled markers are material (\S\ref{sec:method}).}
\label{fig:configs}
\end{figure}

Pooling is positive on 18 of 20 network configurations (Figure~\ref{fig:configs}). The row-shuffled term sits near zero except on the tidal networks. The learned weighting (Q/K) contributes positively to 8 of 10 MV configurations, though materially only on the tidal networks (3 of 10); on ICL configurations it is negative on 9 of 10. The same eight Atlantic tide gauges are worth $+8.15$ points as a multivariate group and $-1.85$ as an in-context one, an effect exacerbated at the longer protocol ($+8.37$ against $-3.81$). Appendix~\ref{app:configs} adds similar results for LSTF corpora and the relatedness controls.

Pooling on those same in-context groups is strongly positive, so the variates carry usable shared content, and it is the identity-dependent part of the weighting that damages the forecast. On 4 of the 10 ICL configurations the full mechanism ends up worse than not attending across variates at all, which is true of none of the 10 multivariate ones (Table~\ref{tab:allconfigs}).

\subsection{Harm is a property of the stack, and block 0 is where to break it}
\label{sec:depth}

\begin{figure}[t]
\centering
\includegraphics[width=\textwidth]{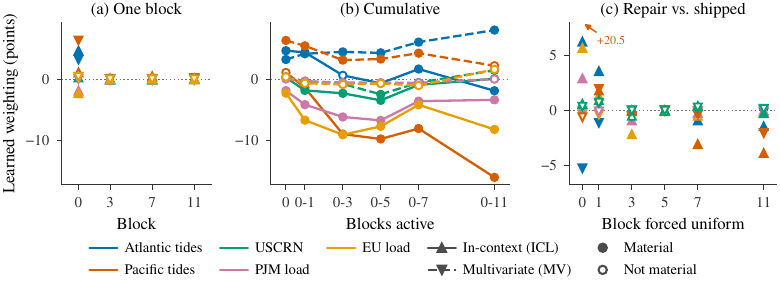}
\caption{Where the learned weighting's cost lives, at the ctx-204 protocol. Filled markers are material (\S\ref{sec:method}); marker direction carries the group type. (a)~one block active, $\alpha = I$ elsewhere; (b)~the leading blocks active; (c)~one block forced uniform (repair), positive meaning better than the unmodified model. Panels (a,b) share a $y$ axis.}
\label{fig:blocks}
\end{figure}

\begin{table}[t]
\centering
\footnotesize
\setlength{\tabcolsep}{3pt}
\caption{Means over the ten sensor-network configurations of each group type, in points of univariate MAE; parentheses count configurations materially positive and negative. Unmodified and repaired models are measured against univariate inference; repair gain is measured against the unmodified model.}
\label{tab:summary}
\begin{tabular}{lrrrrrc}
\toprule
 & & \multicolumn{2}{c}{vs univariate} & \multicolumn{3}{c}{block-0 repair gain} \\ \cmidrule(lr){3-4} \cmidrule(l){5-7}
Group & Learned & Unmodified & Repaired & MAE & WQL & coverage \\
\midrule
MV & $+2.87$ ($3^{+}/0^{-}$) & $+7.10$ ($8^{+}/0^{-}$) & $+5.44$ ($10^{+}/0^{-}$) & $-1.66$ ($0^{+}/3^{-}$) & $-1.59$ & $0.729 \to 0.737$ \\
ICL & $-7.41$ ($0^{+}/8^{-}$) & $-0.74$ ($4^{+}/4^{-}$) & $+6.47$ ($5^{+}/0^{-}$) & $+7.20$ ($8^{+}/0^{-}$) & $+7.59$ & $0.716 \to 0.770$ \\
\midrule
ICL, phase-locked & $-0.79$ ($1^{+}/2^{-}$) & $+5.59$ ($7^{+}/0^{-}$) & $+8.19$ ($10^{+}/0^{-}$) & $+2.60$ ($6^{+}/0^{-}$) & $+2.68$ & $0.757 \to 0.772$ \\
\bottomrule
\end{tabular}
\end{table}

Group attention runs in all twelve blocks, so the effect may depend on where it acts or how many times. We examine where the harm arises by activating group attention in a subset of blocks with $\alpha = I$ elsewhere, and term this procedure \emph{restriction}. In Figure~\ref{fig:blocks}a,b, we find that harm belongs to the stack rather than to any one block. Only block 0 does anything of consequence \emph{individually}, while no other single block moves the forecast by more than $0.52$ points either way. Depth turns block 0's small, mixed-sign effect into a large one: the cumulative ICL subsets deepen the negative effect while the MV side barely moves --- materially negative in 18 of 25 ICL cases against one of 25 for MV (Table~\ref{tab:depth}).

We repair the model by forcing one block to uniform pooling with the rest left learned, and compare against the released Chronos-2. Applied to block 0, this improves all ten in-context configurations across both protocols --- materially on eight (Table~\ref{tab:repair}). On average it gains $+7.20$ points of MAE and $+7.59$ of WQL over the unmodified model, and moves nominal-80\% coverage from $0.716$ to $0.770$, so point accuracy, probabilistic accuracy and calibration improve together rather than trading off (Table~\ref{tab:summary}). On MV it never materially helps in MAE, and the configurations it materially hurts, all tidal, are exactly those where learned weighting materially helped (\S\ref{sec:dissociation}).

Repair pays only in the first two blocks (Table~\ref{tab:repairscan}), while repairing a later block instead \emph{hurts} (Figure~\ref{fig:blocks}c). This fits a chain rather than a single block: block 0's weighting shapes a representation the later blocks keep re-weighting, so on in-context groups it is the composition across depth that costs. Block 0 is thus not the guilty block but the cheapest place to break the chain. Forcing every block uniform gains less than block 0 alone ($+5.92$ against $+7.20$, Table~\ref{tab:repair}), so the weighting is not harmful throughout.

\subsection{Phase dispersion drives the in-context harm}
\label{sec:phase}

% Numbers: results/phaselock_summary.json (scripts/analysis/analyze_phaselock.py).
Our in-context groups' windows start at arbitrary times, so the members' dominant cycles sit at arbitrary phases. Since group attention compares variates at a fixed patch index, we test whether that dispersion causes the harm. For every in-context configuration we build 300 \emph{phase-locked} groups, whose windows are offset from one another by whole periods of the dominant cycle.

Locking phase removes most of the harm (Table~\ref{tab:summary}). The learned weighting becomes materially negative on only 2 of 10 configurations rather than 8, and no phase-locked configuration is worse than univariate inference. Pacific tides are the exception: the harm shrinks but persists ($-16.04 \to -4.92$ and $-14.48 \to -10.31$). Those gauges have mixed semidiurnal and diurnal tides, so locking to the dominant tide may leave the diurnal components dispersed; we have not tested this. Pooling changes little on average, so phase acts mainly through the learned weighting. Alignment does not, however, explain where the weighting \emph{pays}: the tidal multivariate groups, where it helps most, are the least aligned of the multivariate groups (phase alignment $m/N$ of $0.55$--$0.72$, Table~\ref{tab:allconfigs}).

Aligning the examples recovers most of what editing the model buys. Phase-locked groups gain $+5.59$ points over univariate inference with the model unmodified, against $+6.47$ for the original groups after the block-0 repair of \S\ref{sec:depth}. Together the two reach $+8.19$. A jittered control for this experiment is included in Appendix~\ref{app:phase}.

\section{Discussion \& Limitations}
\label{sec:discussion}

Our findings have implications for practitioners: supplying example series, in the style of in-context learning, can be highly effective, provided either the examples are phase-aligned with one another or the learned weighting is removed from block 0 (\S\ref{sec:depth}). Doing both is better than either.
Why the tidal networks are an exception where learned weighting pays, we cannot say. Our edits localize an effect to a pathway and a block without showing what that block computes. Our phase-locking uses one period per network, and the Pacific exception suggests a single period does not capture alignment for signals with several strong cycles. Given all of our interventions occur at inference time, and successive layers consume the output of previous layers, it is unclear whether our findings about depth, position, and repetition would hold if the model were trained from scratch with this modified structure. Notably, Chronos-2 acquires its multivariate ability from synthetic structure imposed on univariate series \citep{chronos2,tirex2}, a choice known to change zero-shot behavior for this class \citep{rmisc}, so what we report may be partly a statement about the training distribution.

\bibliographystyle{unsrtnat}
\bibliography{references}

\appendix
% Appendix floats are numbered A1, A2, ... so a reference to "Table A1" in the main
% text is self-identifying as supplementary rather than something the reader missed.
\setcounter{table}{0}
\setcounter{figure}{0}
\renewcommand{\thetable}{A\arabic{table}}
\renewcommand{\thefigure}{A\arabic{figure}}

\section{All configurations}
\label{app:configs}

Table~\ref{tab:allconfigs} gives every configuration behind the aggregate counts quoted in \S\ref{sec:dissociation}. The sensor-network rows are the data plotted in Figure~\ref{fig:configs}; the LSTF corpora and the relatedness controls (variate sets chosen to be maximally correlated or anti-correlated with the target) are shown numerically only.

\begin{table}[t]
\centering
\small
\caption{All 39 configurations. Contributions in points of univariate-baseline MAE; the three terms add to the total. $^{*}$ marks a material effect: a 95\% paired bootstrap interval excluding $\pm 0.1$ points. $n$ is the number of forecast windows; ctx is context length. $m/N$ is the group's phase alignment: each member's phase is estimated at the dataset's dominant period over the context, and $m/N$ averages $\max(0, \cos(\phi_i - \phi_j))$ over all ordered pairs including $i = j$, so it is $1$ when all members are in phase and about $0.40$ when eight phases are spread uniformly. The sensor-network rows are the data plotted in Figure~\ref{fig:configs}.}
\label{tab:allconfigs}
\begin{tabular}{llrrrrrrr}
\toprule
Dataset & Group & $n$ & ctx & $m/N$ & Pooling & Row-shuffled & Learned & Total \\
\midrule
\multicolumn{9}{l}{\emph{Sensor networks}} \\
Atlantic tides & MV & 273 & 204 & 0.72 & $+5.16^{*}$ & $+1.15^{*}$ & $+8.15^{*}$ & $+14.47^{*}$ \\
Atlantic tides & ICL & 300 & 204 & 0.41 & $+7.59^{*}$ & $+2.00^{*}$ & $-1.85^{*}$ & $+7.74^{*}$ \\
Atlantic tides & MV & 168 & 320 & 0.72 & $+5.41^{*}$ & $+1.19^{*}$ & $+8.37^{*}$ & $+14.97^{*}$ \\
Atlantic tides & ICL & 300 & 320 & 0.41 & $+10.89^{*}$ & $+2.72^{*}$ & $-3.81^{*}$ & $+9.80^{*}$ \\
Pacific tides & MV & 200 & 204 & 0.55 & $+5.26^{*}$ & $+2.82^{*}$ & $+2.31\phantom{^{*}}$ & $+10.40^{*}$ \\
Pacific tides & ICL & 200 & 204 & 0.40 & $+13.77^{*}$ & $+4.14^{*}$ & $-16.04^{*}$ & $+1.87\phantom{^{*}}$ \\
Pacific tides & MV & 168 & 320 & 0.55 & $+5.47^{*}$ & $+3.44^{*}$ & $+6.93^{*}$ & $+15.84^{*}$ \\
Pacific tides & ICL & 200 & 320 & 0.41 & $+14.10^{*}$ & $+4.01^{*}$ & $-14.48^{*}$ & $+3.63^{*}$ \\
USCRN & MV & 200 & 204 & 0.86 & $+1.79^{*}$ & $+0.04\phantom{^{*}}$ & $+1.52\phantom{^{*}}$ & $+3.35^{*}$ \\
USCRN & ICL & 200 & 204 & 0.41 & $+1.71^{*}$ & $+0.30\phantom{^{*}}$ & $+0.14\phantom{^{*}}$ & $+2.15^{*}$ \\
USCRN & MV & 200 & 320 & 0.90 & $+1.79^{*}$ & $-0.09\phantom{^{*}}$ & $-0.34\phantom{^{*}}$ & $+1.37\phantom{^{*}}$ \\
USCRN & ICL & 200 & 320 & 0.42 & $+1.03^{*}$ & $-0.02\phantom{^{*}}$ & $-0.46\phantom{^{*}}$ & $+0.55\phantom{^{*}}$ \\
PJM load & MV & 177 & 204 & 0.89 & $+2.13^{*}$ & $-0.15\phantom{^{*}}$ & $+0.09\phantom{^{*}}$ & $+2.07^{*}$ \\
PJM load & ICL & 200 & 204 & 0.40 & $-0.87\phantom{^{*}}$ & $+0.42\phantom{^{*}}$ & $-3.32^{*}$ & $-3.77^{*}$ \\
PJM load & MV & 108 & 320 & 0.91 & $+1.41^{*}$ & $-0.16\phantom{^{*}}$ & $-0.18\phantom{^{*}}$ & $+1.07\phantom{^{*}}$ \\
PJM load & ICL & 200 & 320 & 0.40 & $+0.41\phantom{^{*}}$ & $+0.18\phantom{^{*}}$ & $-7.52^{*}$ & $-6.94^{*}$ \\
EU load & MV & 63 & 204 & 0.98 & $+1.64^{*}$ & $+0.03\phantom{^{*}}$ & $+1.66\phantom{^{*}}$ & $+3.33^{*}$ \\
EU load & ICL & 200 & 204 & 0.39 & $-0.05\phantom{^{*}}$ & $+1.19^{*}$ & $-8.17^{*}$ & $-7.02^{*}$ \\
EU load & MV & 39 & 320 & 0.98 & $+4.30^{*}$ & $-0.31\phantom{^{*}}$ & $+0.17\phantom{^{*}}$ & $+4.15^{*}$ \\
EU load & ICL & 200 & 320 & 0.39 & $+3.20^{*}$ & $-0.00\phantom{^{*}}$ & $-18.59^{*}$ & $-15.39^{*}$ \\
\midrule
\multicolumn{9}{l}{\emph{LSTF corpora}} \\
ETTh1 & ICL & 400 & 320 & 0.40 & $+1.20^{*}$ & $+0.20\phantom{^{*}}$ & $-1.09^{*}$ & $+0.31\phantom{^{*}}$ \\
ETTh2 & ICL & 381 & 320 & 0.40 & $+0.07\phantom{^{*}}$ & $-1.28^{*}$ & $-1.04\phantom{^{*}}$ & $-2.25^{*}$ \\
ETTm1 & ICL & 400 & 320 & 0.40 & $+1.36^{*}$ & $+0.08\phantom{^{*}}$ & $+0.90\phantom{^{*}}$ & $+2.34^{*}$ \\
electricity & ICL & 397 & 320 & 0.40 & $-0.61\phantom{^{*}}$ & $-0.32\phantom{^{*}}$ & $-14.08^{*}$ & $-15.00^{*}$ \\
exchange\_rate & ICL & 400 & 320 & 0.41 & $+2.08^{*}$ & $+0.73^{*}$ & $+0.51\phantom{^{*}}$ & $+3.32^{*}$ \\
solar\_AL & ICL & 400 & 320 & 0.40 & $-0.88^{*}$ & $+0.37^{*}$ & $-0.86^{*}$ & $-1.37^{*}$ \\
traffic & ICL & 399 & 320 & 0.40 & $-0.70^{*}$ & $-0.06\phantom{^{*}}$ & $-5.45^{*}$ & $-6.22^{*}$ \\
ETTh1 & MV & 41 & 320 & 0.46 & $+1.35^{*}$ & $-0.03\phantom{^{*}}$ & $+2.35\phantom{^{*}}$ & $+3.67\phantom{^{*}}$ \\
ETTh2 & MV & 39 & 320 & 0.60 & $+0.34\phantom{^{*}}$ & $-0.54\phantom{^{*}}$ & $-0.16\phantom{^{*}}$ & $-0.36\phantom{^{*}}$ \\
ETTm1 & MV & 167 & 320 & 0.49 & $+1.28^{*}$ & $+0.36\phantom{^{*}}$ & $+0.64\phantom{^{*}}$ & $+2.28\phantom{^{*}}$ \\
electricity & MV & 397 & 320 & 0.75 & $+1.10\phantom{^{*}}$ & $+0.13\phantom{^{*}}$ & $-0.83\phantom{^{*}}$ & $+0.40\phantom{^{*}}$ \\
solar\_AL & MV & 400 & 320 & 0.94 & $+0.35^{*}$ & $-0.01\phantom{^{*}}$ & $+0.16\phantom{^{*}}$ & $+0.51^{*}$ \\
traffic & MV & 400 & 320 & 0.82 & $+0.78^{*}$ & $-0.24^{*}$ & $-0.82^{*}$ & $-0.28\phantom{^{*}}$ \\
\midrule
\multicolumn{9}{l}{\emph{Relatedness controls}} \\
electricity (anti) & MV & 87 & 320 & 0.48 & $-1.68\phantom{^{*}}$ & $-1.35\phantom{^{*}}$ & $-4.43^{*}$ & $-7.47\phantom{^{*}}$ \\
solar\_AL (anti) & MV & 150 & 320 & 0.93 & $+0.59^{*}$ & $-0.03\phantom{^{*}}$ & $-0.02\phantom{^{*}}$ & $+0.53^{*}$ \\
traffic (anti) & MV & 150 & 320 & 0.69 & $+0.41\phantom{^{*}}$ & $-0.13\phantom{^{*}}$ & $+0.07\phantom{^{*}}$ & $+0.34\phantom{^{*}}$ \\
electricity (corr) & MV & 148 & 320 & 0.96 & $+0.89\phantom{^{*}}$ & $-0.19\phantom{^{*}}$ & $-2.51\phantom{^{*}}$ & $-1.80\phantom{^{*}}$ \\
solar\_AL (corr) & MV & 150 & 320 & 0.98 & $+0.57^{*}$ & $-0.05\phantom{^{*}}$ & $+0.02\phantom{^{*}}$ & $+0.54^{*}$ \\
traffic (corr) & MV & 150 & 320 & 0.98 & $+0.47\phantom{^{*}}$ & $-0.27\phantom{^{*}}$ & $-0.23\phantom{^{*}}$ & $-0.03\phantom{^{*}}$ \\
\bottomrule
\end{tabular}
\end{table}

\section{Controls}
\label{app:controls}

Table~\ref{tab:controls} tests whether the learned V/O projection can be replaced by a raw or norm-matched cross-variate mean.

\begin{table}[t]
\centering
\small
\caption{Is the V/O projection necessary? Each column is that condition's error relative to the univariate baseline, in points. \textsc{raw mean} substitutes the unprojected cross-variate mean for $W_{OV}$; \textsc{matched} rescales that mean to the norm of the learned update; \textsc{uniform} and \textsc{learned} are the settings of \S\ref{sec:method}. Replacing the projection costs far more than the entire effect the paper reports, so the uniform-$\alpha$ result is not a statement that cross-variate attention reduces to averaging.}
\label{tab:controls}
\begin{tabular}{llrrrrr}
\toprule
 & & & \multicolumn{4}{c}{error vs univariate (points)} \\
\cmidrule(l){4-7}
Configuration & Group & $n$ & raw mean & matched & uniform & learned \\
\midrule
\texttt{tidesp256\_icl} & ICL & 250 & $-216.8$ & $-14.1$ & $+7.71$ & $+7.17$ \\
\texttt{netpjmp256\_icl} & ICL & 200 & $-98.3$ & $-8.7$ & $-0.87$ & $-3.77$ \\
\texttt{tidesp256\_mv} & MV & 250 & $-222.1$ & $-51.0$ & $+5.18$ & $+14.59$ \\
\texttt{netuscrnp256\_mv} & MV & 200 & $-63.6$ & $-9.3$ & $+1.79$ & $+3.35$ \\
\bottomrule
\end{tabular}
\end{table}

\section{Phase-locked groups}
\label{app:phase}

Each phase-locked group takes one station and eight window start times $t_k = t_0 + n_k P$, where $P$ is the network's dominant period (the M2 tide, $12.42$\,h, for both tidal networks; $24$\,h for USCRN; $168$\,h for the two load networks, which locks the daily cycle as well), $t_0$ is drawn uniformly in $[0, P)$ per group so forecasts are not all issued at one time of day, and the integers $n_k$ keep windows at least one window length plus $P$ apart, so no window's context contains another's horizon. PJM is assembled on parsed timestamps, since its zone intersection has gaps. The paired jittered control shifts each $t_k$ by an independent offset drawn uniformly in $[0, P)$, restoring dispersion ($m/N \approx 0.40$) while keeping station, anchors and spacing. It reproduces the original in-context ladder: the intervals of 39 of its 40 terms overlap the original's (the exception is pooling on EU load at ctx-320, the shortest record), its learned weighting averages $-7.75$ against $-7.41$, and univariate error differs between locked and jittered groups by at most $6\%$, so the arms do not differ in difficulty.

\section{The restriction sweep}
\label{app:depth}

Table~\ref{tab:depth} gives every condition behind the restriction counts in \S\ref{sec:depth}, and Table~\ref{tab:depthpool} the pooling term for the same sweep. Both are at the ctx-204 protocol.

\begin{table}[t]
\centering
\footnotesize
\setlength{\tabcolsep}{2.8pt}
\caption{The restriction sweep. Group attention is active only in the named blocks, with $\alpha = I$ in the rest, and the full ladder of \S\ref{sec:method} is run inside that restricted model; each entry is its learned-weighting term (learned minus row-shuffled), in points of univariate-baseline MAE. It is not the total gain over univariate inference: the `all' column re-measures the Learned column of Table~\ref{tab:allconfigs} (e.g.\ Atlantic tides ICL $-1.84$ here, $-1.85$ there), not its Total. $^{*}$ marks a material effect (\S\ref{sec:method}). Left group: one block in isolation. Middle: the leading blocks together. Right: the last four blocks only. In isolation no block other than 0 moves the forecast by more than $0.52$ points either way, while block 0 spans $-2.23$ to $+6.44$. Of the 5 multi-block subsets on 5 in-context datasets, 18 of 25 are materially negative and the worst reaches $-16.09$; on multivariate datasets 1 of 25 are, and the worst is $-2.43$. With only the last four blocks active, no dataset moves by more than $0.88$ points.}
\label{tab:depth}
\begin{tabular}{llrrrrrrrrrr}
\toprule
 & & \multicolumn{4}{c}{one block} & \multicolumn{5}{c}{leading blocks} & \multicolumn{1}{c}{late} \\
\cmidrule(lr){3-6} \cmidrule(lr){7-11} \cmidrule(l){12-12}
Dataset & Group & 0 & 3 & 7 & 11 & 0--1 & 0--3 & 0--5 & 0--7 & all & 8--11 \\
\midrule
Atlantic tides & ICL & $+4.75^{*}$ & $+0.06\phantom{^{*}}$ & $+0.12\phantom{^{*}}$ & $+0.07\phantom{^{*}}$ & $+4.49^{*}$ & $+0.70\phantom{^{*}}$ & $-0.61\phantom{^{*}}$ & $+1.74^{*}$ & $-1.84^{*}$ & $+0.15\phantom{^{*}}$ \\
Pacific tides & ICL & $+1.18\phantom{^{*}}$ & $+0.15^{*}$ & $+0.52^{*}$ & $+0.22^{*}$ & $-1.34\phantom{^{*}}$ & $-8.97^{*}$ & $-9.77^{*}$ & $-8.05^{*}$ & $-16.09^{*}$ & $+0.88^{*}$ \\
USCRN & ICL & $+0.34\phantom{^{*}}$ & $-0.01\phantom{^{*}}$ & $-0.01\phantom{^{*}}$ & $+0.08\phantom{^{*}}$ & $-1.78^{*}$ & $-2.26^{*}$ & $-3.41^{*}$ & $-0.86\phantom{^{*}}$ & $+0.14\phantom{^{*}}$ & $+0.10\phantom{^{*}}$ \\
PJM load & ICL & $-1.82^{*}$ & $+0.00\phantom{^{*}}$ & $+0.01\phantom{^{*}}$ & $+0.04\phantom{^{*}}$ & $-4.12^{*}$ & $-6.15^{*}$ & $-6.71^{*}$ & $-3.55^{*}$ & $-3.34^{*}$ & $+0.02\phantom{^{*}}$ \\
EU load & ICL & $-2.23^{*}$ & $+0.06\phantom{^{*}}$ & $+0.13\phantom{^{*}}$ & $+0.11\phantom{^{*}}$ & $-6.67^{*}$ & $-9.08^{*}$ & $-7.70^{*}$ & $-4.14^{*}$ & $-8.20^{*}$ & $+0.68^{*}$ \\
Atlantic tides & MV & $+3.30^{*}$ & $+0.06\phantom{^{*}}$ & $+0.10\phantom{^{*}}$ & $+0.13\phantom{^{*}}$ & $+4.25^{*}$ & $+4.55^{*}$ & $+4.40^{*}$ & $+6.16^{*}$ & $+8.14^{*}$ & $+0.45^{*}$ \\
Pacific tides & MV & $+6.44^{*}$ & $+0.12\phantom{^{*}}$ & $+0.13^{*}$ & $+0.26\phantom{^{*}}$ & $+5.55^{*}$ & $+3.21^{*}$ & $+3.38^{*}$ & $+4.35^{*}$ & $+2.28\phantom{^{*}}$ & $+0.55^{*}$ \\
USCRN & MV & $+0.47^{*}$ & $+0.01\phantom{^{*}}$ & $-0.02\phantom{^{*}}$ & $+0.00\phantom{^{*}}$ & $-0.26\phantom{^{*}}$ & $-0.71\phantom{^{*}}$ & $-2.43^{*}$ & $-0.49\phantom{^{*}}$ & $+1.53\phantom{^{*}}$ & $+0.07\phantom{^{*}}$ \\
PJM load & MV & $+0.11\phantom{^{*}}$ & $+0.00\phantom{^{*}}$ & $+0.02\phantom{^{*}}$ & $-0.03\phantom{^{*}}$ & $-0.25\phantom{^{*}}$ & $-0.49\phantom{^{*}}$ & $-0.51\phantom{^{*}}$ & $-0.53\phantom{^{*}}$ & $+0.09\phantom{^{*}}$ & $-0.15\phantom{^{*}}$ \\
EU load & MV & $+0.44\phantom{^{*}}$ & $+0.03\phantom{^{*}}$ & $+0.05\phantom{^{*}}$ & $+0.01\phantom{^{*}}$ & $-0.60\phantom{^{*}}$ & $-0.80\phantom{^{*}}$ & $-0.64\phantom{^{*}}$ & $-0.98\phantom{^{*}}$ & $+1.68\phantom{^{*}}$ & $+0.36^{*}$ \\
\bottomrule
\end{tabular}
\end{table}

\begin{table}[t]
\centering
\small
\caption{The other half of the restriction sweep: pooling (the V/O pathway) with one block active against all twelve, in points of univariate-baseline MAE. $^{*}$ marks a material effect. Pooling grows with the number of active blocks on 8 of 10 datasets, the opposite of the learned weighting in Table~\ref{tab:depth}.}
\label{tab:depthpool}
\begin{tabular}{llrr}
\toprule
Dataset & Group & block 0 only & all twelve \\
\midrule
Atlantic tides & ICL & $+0.20\phantom{^{*}}$ & $+7.59^{*}$ \\
Pacific tides & ICL & $+1.14^{*}$ & $+13.77^{*}$ \\
USCRN & ICL & $-0.04\phantom{^{*}}$ & $+1.71^{*}$ \\
PJM load & ICL & $-0.06\phantom{^{*}}$ & $-0.87\phantom{^{*}}$ \\
EU load & ICL & $+0.10\phantom{^{*}}$ & $-0.05\phantom{^{*}}$ \\
Atlantic tides & MV & $+0.21^{*}$ & $+5.16^{*}$ \\
Pacific tides & MV & $+0.58^{*}$ & $+5.26^{*}$ \\
USCRN & MV & $+0.10\phantom{^{*}}$ & $+1.79^{*}$ \\
PJM load & MV & $+0.13\phantom{^{*}}$ & $+2.13^{*}$ \\
EU load & MV & $+0.24\phantom{^{*}}$ & $+1.64^{*}$ \\
\bottomrule
\end{tabular}
\end{table}

\section{The repair intervention}
\label{app:repair}

Table~\ref{tab:repair} gives every configuration behind the repair result in \S\ref{sec:depth}, and Table~\ref{tab:repairscan} the single-block scan that localizes it. The UV MAE column is the univariate baseline every contribution in this paper is measured against, in the native units of each dataset.

\begin{table}[t]
\centering
\small
\caption{The repair intervention on every sensor-network configuration: $\alpha$ forced uniform in the named blocks of the otherwise-shipped model, against the shipped model itself. Positive is better than shipped, in points of univariate-baseline MAE (or of the univariate WQL). $^{*}$ marks a material effect: a 95\% paired bootstrap interval excluding $\pm 0.1$ points, the floor calibrated by the block-5 near-null. Coverage is the empirical coverage of the nominal $80\%$ interval, shipped $\to$ repaired. Repairing block 0 alone improves every in-context configuration on all three metrics; its only material costs are on tidal multivariate groups, which are also the only configurations where learned weighting materially helps (Table~\ref{tab:allconfigs}).}
\label{tab:repair}
\begin{tabular}{llrrrrrcr}
\toprule
 & & & & & \multicolumn{3}{c}{block 0 uniform} & all blocks \\
\cmidrule(l){6-8} \cmidrule(l){9-9}
Dataset & Group & ctx & $n$ & UV MAE & MAE & WQL & coverage & MAE \\
\midrule
\multicolumn{9}{l}{\emph{In-context groups}} \\
Atlantic tides & ICL & 204 & 300 & $0.1404$ & $+6.27^{*}$ & $+5.56^{*}$ & $0.751 \to 0.758$ & $-0.15\phantom{^{*}}$ \\
Atlantic tides & ICL & 320 & 300 & $0.1705$ & $+6.45^{*}$ & $+6.58^{*}$ & $0.700 \to 0.776$ & $+1.09\phantom{^{*}}$ \\
Pacific tides & ICL & 204 & 200 & $0.1228$ & $+20.49^{*}$ & $+19.53^{*}$ & $0.677 \to 0.827$ & $+11.90^{*}$ \\
Pacific tides & ICL & 320 & 200 & $0.1486$ & $+10.85^{*}$ & $+13.38^{*}$ & $0.593 \to 0.708$ & $+10.46^{*}$ \\
USCRN & ICL & 204 & 200 & $3.203$ & $+0.58\phantom{^{*}}$ & $+0.56\phantom{^{*}}$ & $0.760 \to 0.772$ & $-0.44\phantom{^{*}}$ \\
USCRN & ICL & 320 & 200 & $3.585$ & $+0.40\phantom{^{*}}$ & $+0.89^{*}$ & $0.746 \to 0.769$ & $+0.48\phantom{^{*}}$ \\
PJM load & ICL & 204 & 200 & $442.5$ & $+2.92^{*}$ & $+3.27^{*}$ & $0.744 \to 0.772$ & $+2.90^{*}$ \\
PJM load & ICL & 320 & 200 & $483.6$ & $+5.63^{*}$ & $+8.15^{*}$ & $0.699 \to 0.777$ & $+7.35^{*}$ \\
EU load & ICL & 204 & 200 & $867.5$ & $+5.69^{*}$ & $+8.67^{*}$ & $0.716 \to 0.754$ & $+6.98^{*}$ \\
EU load & ICL & 320 & 200 & $864.5$ & $+12.74^{*}$ & $+9.35^{*}$ & $0.774 \to 0.791$ & $+18.59^{*}$ \\
\midrule
\multicolumn{9}{l}{\emph{Multivariate groups}} \\
Atlantic tides & MV & 204 & 273 & $0.1402$ & $-5.30^{*}$ & $-5.97^{*}$ & $0.710 \to 0.668$ & $-9.30^{*}$ \\
Atlantic tides & MV & 320 & 168 & $0.1755$ & $-5.88^{*}$ & $-6.42^{*}$ & $0.713 \to 0.700$ & $-9.56^{*}$ \\
Pacific tides & MV & 204 & 200 & $0.1169$ & $-0.64\phantom{^{*}}$ & $-1.27\phantom{^{*}}$ & $0.737 \to 0.728$ & $-5.13^{*}$ \\
Pacific tides & MV & 320 & 168 & $0.1457$ & $-6.13^{*}$ & $-4.20^{*}$ & $0.595 \to 0.706$ & $-10.37^{*}$ \\
USCRN & MV & 204 & 200 & $3.180$ & $+0.32\phantom{^{*}}$ & $+0.38\phantom{^{*}}$ & $0.763 \to 0.773$ & $-1.56\phantom{^{*}}$ \\
USCRN & MV & 320 & 200 & $3.509$ & $+0.71\phantom{^{*}}$ & $+1.31^{*}$ & $0.751 \to 0.767$ & $+0.43\phantom{^{*}}$ \\
PJM load & MV & 204 & 177 & $466.8$ & $-0.08\phantom{^{*}}$ & $-0.04\phantom{^{*}}$ & $0.719 \to 0.724$ & $+0.06\phantom{^{*}}$ \\
PJM load & MV & 320 & 108 & $508.3$ & $+0.54\phantom{^{*}}$ & $+0.50\phantom{^{*}}$ & $0.753 \to 0.759$ & $+0.33\phantom{^{*}}$ \\
EU load & MV & 204 & 63 & $831.2$ & $-0.37\phantom{^{*}}$ & $-0.05\phantom{^{*}}$ & $0.765 \to 0.761$ & $-1.69\phantom{^{*}}$ \\
EU load & MV & 320 & 39 & $805.9$ & $+0.24\phantom{^{*}}$ & $-0.09\phantom{^{*}}$ & $0.782 \to 0.779$ & $+0.15\phantom{^{*}}$ \\
\bottomrule
\end{tabular}
\end{table}

\begin{table}[t]
\centering
\footnotesize
\setlength{\tabcolsep}{2.8pt}
\caption{Which block has to be neutralized, on all 20 configurations. Each column forces $\alpha$ uniform in the named block or blocks of the otherwise-shipped model; positive is better than shipped, in points of univariate-baseline MAE. $^{*}$ marks a 95\% paired bootstrap interval excluding $\pm 0.1$ points, the floor this column calibrates. Nearly all of the recoverable effect sits in blocks 0 and 1, and no single later block recovers more than $0.39$ points. Block 5 is the intended null: its attention is already near-uniform in the shipped model, so forcing it uniform moves the forecast by at most $0.081$ points, which is material on none of the 20 --- though 14 of the 20 intervals do exclude zero, which is what calibrates the floor and why effects are reported against it rather than against zero.}
\label{tab:repairscan}
\begin{tabular}{llrrrrrrrrrr}
\toprule
 & & & \multicolumn{9}{c}{blocks forced uniform} \\
\cmidrule(l){4-12}
Dataset & Group & ctx & 0 & 1 & 3 & 5 & 7 & 11 & 0--1 & 0--3 & all \\
\midrule
Atlantic tides & ICL & 204 & $+6.27^{*}$ & $+3.57^{*}$ & $-0.00\phantom{^{*}}$ & $-0.01\phantom{^{*}}$ & $-0.94^{*}$ & $-1.46^{*}$ & $+1.57^{*}$ & $+0.62\phantom{^{*}}$ & $-0.15\phantom{^{*}}$ \\
Atlantic tides & ICL & 320 & $+6.45^{*}$ & $+5.72^{*}$ & $+0.02\phantom{^{*}}$ & $-0.00\phantom{^{*}}$ & $-0.31\phantom{^{*}}$ & $-0.64^{*}$ & $+2.89^{*}$ & $+1.95^{*}$ & $+1.09\phantom{^{*}}$ \\
Pacific tides & ICL & 204 & $+20.49^{*}$ & $+1.80^{*}$ & $-0.08\phantom{^{*}}$ & $-0.03\phantom{^{*}}$ & $-3.10^{*}$ & $-3.90^{*}$ & $+15.57^{*}$ & $+14.26^{*}$ & $+11.90^{*}$ \\
Pacific tides & ICL & 320 & $+10.85^{*}$ & $+3.00^{*}$ & $+0.30^{*}$ & $-0.04\phantom{^{*}}$ & $-0.54^{*}$ & $-1.61^{*}$ & $+12.76^{*}$ & $+11.87^{*}$ & $+10.46^{*}$ \\
USCRN & ICL & 204 & $+0.58\phantom{^{*}}$ & $+0.92^{*}$ & $-0.50\phantom{^{*}}$ & $-0.03\phantom{^{*}}$ & $+0.39^{*}$ & $-0.27^{*}$ & $-0.41\phantom{^{*}}$ & $-0.49\phantom{^{*}}$ & $-0.44\phantom{^{*}}$ \\
USCRN & ICL & 320 & $+0.40\phantom{^{*}}$ & $+0.69^{*}$ & $-2.06^{*}$ & $-0.03\phantom{^{*}}$ & $+0.14\phantom{^{*}}$ & $-0.04\phantom{^{*}}$ & $+0.14\phantom{^{*}}$ & $+0.07\phantom{^{*}}$ & $+0.48\phantom{^{*}}$ \\
PJM load & ICL & 204 & $+2.92^{*}$ & $+0.36\phantom{^{*}}$ & $-0.93^{*}$ & $-0.03\phantom{^{*}}$ & $-0.24\phantom{^{*}}$ & $-0.04\phantom{^{*}}$ & $+2.91^{*}$ & $+2.85^{*}$ & $+2.90^{*}$ \\
PJM load & ICL & 320 & $+5.63^{*}$ & $+1.26^{*}$ & $-1.52^{*}$ & $-0.04\phantom{^{*}}$ & $+0.20\phantom{^{*}}$ & $+0.01\phantom{^{*}}$ & $+7.40^{*}$ & $+7.46^{*}$ & $+7.35^{*}$ \\
EU load & ICL & 204 & $+5.69^{*}$ & $+1.32^{*}$ & $-2.19^{*}$ & $-0.08\phantom{^{*}}$ & $-0.49\phantom{^{*}}$ & $-0.19\phantom{^{*}}$ & $+9.35^{*}$ & $+9.43^{*}$ & $+6.98^{*}$ \\
EU load & ICL & 320 & $+12.74^{*}$ & $+3.44^{*}$ & $-3.31^{*}$ & $-0.04\phantom{^{*}}$ & $+0.18\phantom{^{*}}$ & $-0.22^{*}$ & $+18.08^{*}$ & $+18.28^{*}$ & $+18.59^{*}$ \\
Atlantic tides & MV & 204 & $-5.30^{*}$ & $-1.15^{*}$ & $-0.54^{*}$ & $+0.00\phantom{^{*}}$ & $-0.37^{*}$ & $-0.17\phantom{^{*}}$ & $-7.27^{*}$ & $-7.91^{*}$ & $-9.30^{*}$ \\
Atlantic tides & MV & 320 & $-5.88^{*}$ & $-1.45^{*}$ & $-0.09\phantom{^{*}}$ & $+0.01\phantom{^{*}}$ & $-0.20\phantom{^{*}}$ & $-0.12\phantom{^{*}}$ & $-8.32^{*}$ & $-8.73^{*}$ & $-9.56^{*}$ \\
Pacific tides & MV & 204 & $-0.64\phantom{^{*}}$ & $+1.95^{*}$ & $-0.54^{*}$ & $+0.00\phantom{^{*}}$ & $-0.41^{*}$ & $-2.08^{*}$ & $-2.62\phantom{^{*}}$ & $-3.65^{*}$ & $-5.13^{*}$ \\
Pacific tides & MV & 320 & $-6.13^{*}$ & $+1.00^{*}$ & $-0.02\phantom{^{*}}$ & $+0.01\phantom{^{*}}$ & $-0.15\phantom{^{*}}$ & $-1.22^{*}$ & $-8.47^{*}$ & $-9.24^{*}$ & $-10.37^{*}$ \\
USCRN & MV & 204 & $+0.32\phantom{^{*}}$ & $+0.69\phantom{^{*}}$ & $-0.02\phantom{^{*}}$ & $-0.02\phantom{^{*}}$ & $+0.25\phantom{^{*}}$ & $+0.14\phantom{^{*}}$ & $-1.00\phantom{^{*}}$ & $-1.21\phantom{^{*}}$ & $-1.56\phantom{^{*}}$ \\
USCRN & MV & 320 & $+0.71\phantom{^{*}}$ & $+1.52^{*}$ & $-0.76\phantom{^{*}}$ & $-0.01\phantom{^{*}}$ & $+0.17\phantom{^{*}}$ & $+0.17\phantom{^{*}}$ & $+0.43\phantom{^{*}}$ & $+0.38\phantom{^{*}}$ & $+0.43\phantom{^{*}}$ \\
PJM load & MV & 204 & $-0.08\phantom{^{*}}$ & $-0.11\phantom{^{*}}$ & $+0.02\phantom{^{*}}$ & $+0.00\phantom{^{*}}$ & $-0.08\phantom{^{*}}$ & $+0.00\phantom{^{*}}$ & $-0.41\phantom{^{*}}$ & $-0.40\phantom{^{*}}$ & $+0.06\phantom{^{*}}$ \\
PJM load & MV & 320 & $+0.54\phantom{^{*}}$ & $+0.49\phantom{^{*}}$ & $-0.22\phantom{^{*}}$ & $-0.01\phantom{^{*}}$ & $+0.06\phantom{^{*}}$ & $+0.02\phantom{^{*}}$ & $+0.34\phantom{^{*}}$ & $+0.23\phantom{^{*}}$ & $+0.33\phantom{^{*}}$ \\
EU load & MV & 204 & $-0.37\phantom{^{*}}$ & $-0.38\phantom{^{*}}$ & $-0.04\phantom{^{*}}$ & $-0.05\phantom{^{*}}$ & $-0.16\phantom{^{*}}$ & $+0.06\phantom{^{*}}$ & $-0.77\phantom{^{*}}$ & $-1.01\phantom{^{*}}$ & $-1.69\phantom{^{*}}$ \\
EU load & MV & 320 & $+0.24\phantom{^{*}}$ & $-0.07\phantom{^{*}}$ & $-0.06\phantom{^{*}}$ & $-0.04\phantom{^{*}}$ & $-0.09\phantom{^{*}}$ & $+0.05\phantom{^{*}}$ & $-0.18\phantom{^{*}}$ & $-0.45\phantom{^{*}}$ & $+0.15\phantom{^{*}}$ \\
\bottomrule
\end{tabular}
\end{table}

\end{document}